\documentclass[10pt,twocolumn,letterpaper]{article}

\usepackage{wacv}
\usepackage{times}
\usepackage{epsfig}
\usepackage{graphicx}
\usepackage{amsmath}
\usepackage{amssymb}
\usepackage{booktabs}
\usepackage{graphicx}
\usepackage{makecell}
\usepackage{subfig}
\usepackage{amsmath}
\usepackage{algorithm}
\usepackage{algorithmic}
\usepackage{float}
\usepackage{makecell}
\usepackage{bbm}
\usepackage{siunitx}
\usepackage{pdfpages}

\def\wacvPaperID{****} % Enter the WACV Paper ID here

\wacvalgorithmstrack   % Uncomment this line if you are submitting to the Algorithms Track.
\wacvfinalcopy % *** Uncomment this line for the final submission

\ifwacvfinal
\usepackage[breaklinks=true,bookmarks=false]{hyperref}
\else
\usepackage[pagebackref=true,breaklinks=true,colorlinks,bookmarks=false]{hyperref}
\fi

\begin{document}

%%%%%%%%% TITLE
\title{More Than Just Attention: Improving Cross-Modal Attentions \\ with Contrastive Constraints for Image-Text Matching}

% \author{Yuxiao Chen\\
% \and
% Jianbo Yuan\\
% \and
% Long Zhao\\
% \and
% Jianbo Yuan\\

% \and
% Long Zhao\\

% \and
% Tianlang Chen\\

% \and
% Rui Luo\\

% \and
% Larry Davis\\

% \and
% Dimitris N. Metaxas\\

% }
\author{
    Yuxiao Chen\textsuperscript{\rm 1} \thanks{This work was done while Yuxiao Chen was a research intern at Amazon. Correspondence to: Yuxiao Chen~(yc984@cs.rutgers.edu)},
    Jianbo Yuan\textsuperscript{\rm 2},
    Long Zhao\textsuperscript{\rm 1},
    Tianlang Chen\textsuperscript{\rm 2},
    Rui Luo\textsuperscript{\rm 2}, \\
    Larry Davis\textsuperscript{\rm 2},
    Dimitris N. Metaxas\textsuperscript{\rm 1} \\
    \\
    \textsuperscript{\rm 1}Rutgers University,
    \textsuperscript{\rm 2}Amazon.com Services, Inc \\
}
% \{yc984, lz311, dnm\}@cs.rutgers.edu \{yjianbo, ctianlan, luorui, lrrydav\}@amazon.com

    %Afiliations
    % \textsuperscript{\rm 1}Rutgers University
    % \textsuperscript{\rm 2}Amazon.com Services, Inc \\
    % % email address must be in roman text type, not monospace or sans serif
    % \{yc984, lz311, dnm\}@cs.rutgers.edu \{yjianbo, luorui, lrrydav\}@amazon.com
%
% See more examples next

\maketitle
\thispagestyle{empty}

%%%%%%%%% ABSTRACT
\begin{abstract}
Cross-modal attention mechanisms have been widely applied to the image-text matching task. They have achieved remarkable improvements thanks to their capability of learning fine-grained relevance across different modalities. However, the cross-modal attention models of existing methods could be sub-optimal and inaccurate because there is no direct supervision provided during the training process. In this work, we propose two novel training strategies, namely Contrastive Content Re-sourcing (CCR) and Contrastive Content Swapping (CCS) constraints, to address such limitations. These constraints supervise the training of cross-modal attention models in a contrastive learning manner without requiring explicit attention annotations. They are plug-in training strategies and can be generally integrated into existing cross-modal attention models. Additionally, we introduce three metrics, including Attention Precision, Recall, and F1-Score, to quantitatively measure the quality of learned attention models. We evaluate the proposed constraints by incorporating them into four state-of-the-art cross-modal attention-based image-text matching models. Experimental results on both Flickr30k and MS-COCO datasets demonstrate that integrating these constraints generally improves the model performance in terms of both retrieval performance and attention metrics.
\end{abstract}

\begin{figure}[!t]
  \includegraphics[width=\linewidth]{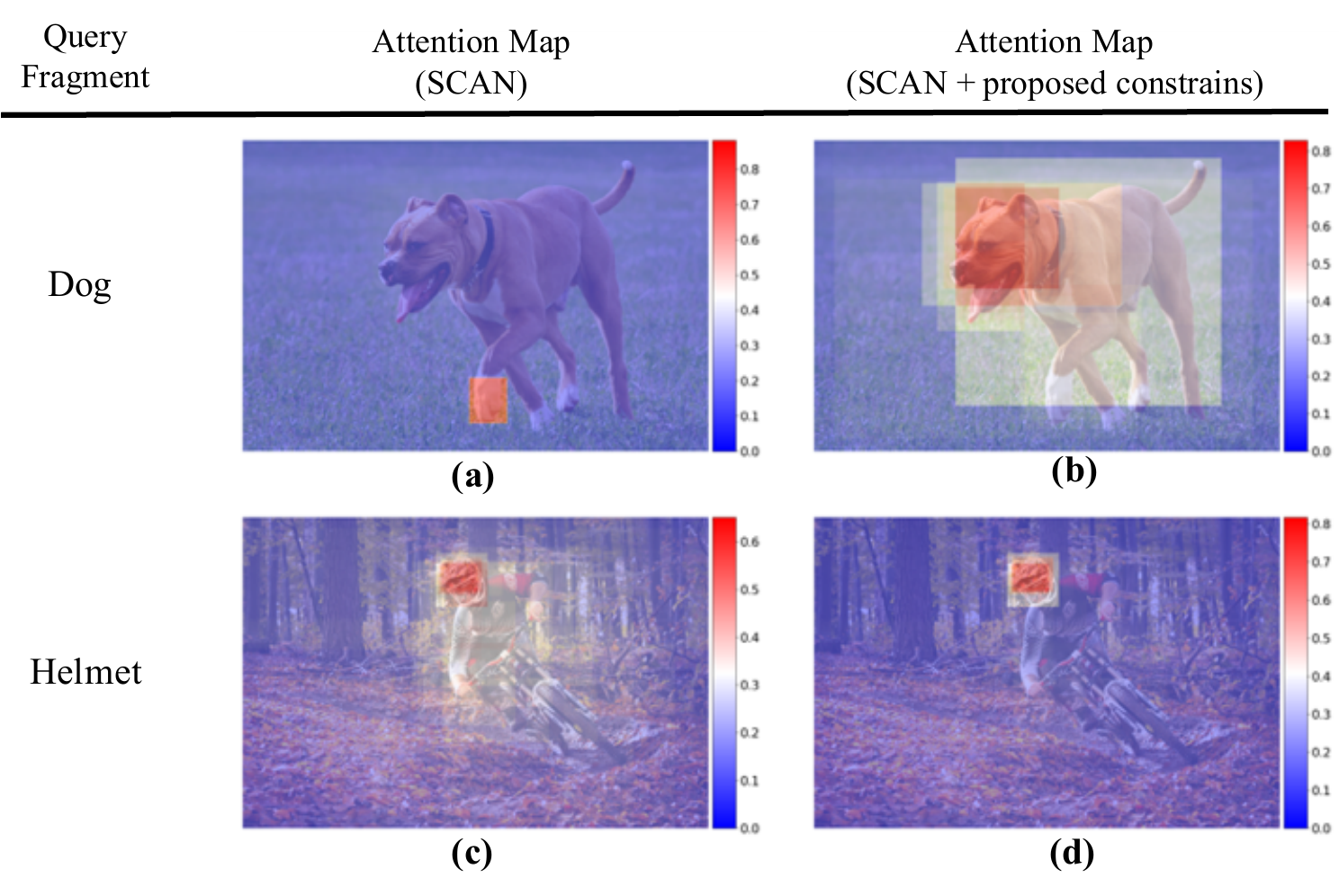}
  \caption{Visualization of the attention maps of the SCAN model learned without and with our proposed constraints.}
  \label{fig:fig_1}
\end{figure}

\begin{figure*}[!t]
    
  \includegraphics[width=\textwidth]{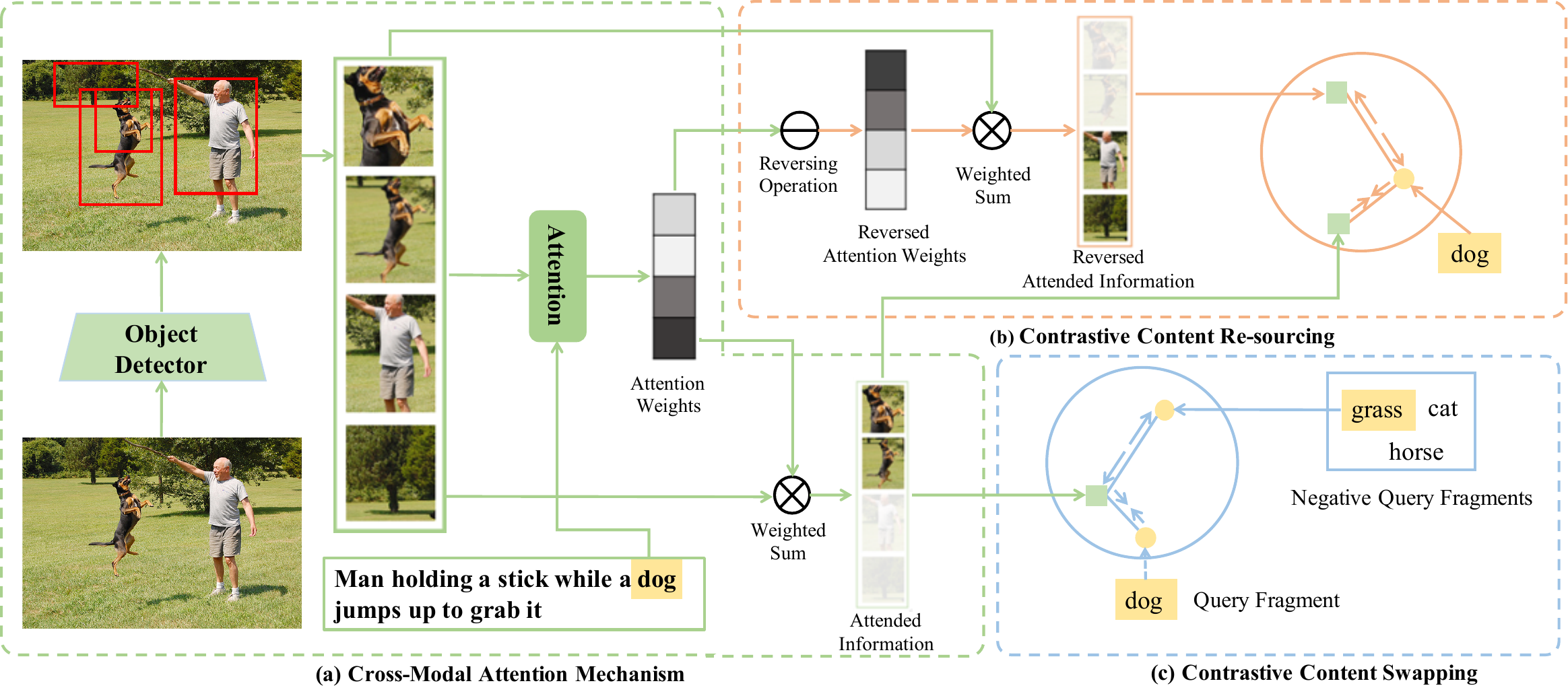}
  \caption{Overview of the training pipeline which contains \textbf{(a)} the cross-modal attention mechanism and our proposed attention constraints including \textbf{(b)} Contrastive Content Re-sourcing~(CCR) and \textbf{(c)} Contrastive Content Swapping~(CCS).} %For CCR (b), the image fragments are divided into two groups based on their attention weights with respect to the word ``dog", and are used as the positive and negative sample of the word ``dog". In CCS (c), the sampled word ``grass" and the query ``dog" is used as the negative and positive sample, respectively.}
  \label{fig:fig_2}
\end{figure*}

%%%%%%%%% BODY TEXT
\section{Introduction}
\label{sec:intro}
The task of image-text matching aims to learn a model that measures the similarity between visual and textual contents. By using the learned model, users can retrieve images that visually match the context described by a text query, or retrieve texts that best describe the image query. Because of its critical role to bridge the human vision and language world, this task has emerged as an active research area~\cite{faghri2017vse++, nam2017dual,huang2017instance,lee2018stacked,liu2019focus,chen2020adaptive, wang2019position}.

% The key idea of existing work is to encode text and image information into a common embedding space. Early approaches implement this by treating each image or sentence as an entity, and then using neural networks to encode sentences and images in to a latent space. However, these methods fail to capture fine-grained correspondence and difference between sentences and images and so achieve sub-optimal performance. Recent studies go to a finer level by embeds fragments of images (objects) and fragments of sentences (words) into a common space. However, they ignore the fact that the importance of fragments varies when context change.

Recently, cross-modal attention models have been widely applied to this task~\cite{liu2019focus,lee2018stacked,nam2017dual,huang2017instance,huang2018bi, chen2019uniter, li2020oscar, diao2021similarity}. These approaches have achieved remarkable improvements thanks to their ability to capture fine-grained cross-modal relevance by the cross-modal attention mechanism. Specifically, given an image description and its corresponding image, they are first represented by fragments, i.e., individual words and image regions. We refer to the fragments of the context modality as query fragments, and the fragments of the attended modality as key fragments. Given a query fragment, a cross-modal attention model first assigns an attention weight to each key fragment, each of which measures the semantic relevance between the query fragment and the corresponding key fragment. Then the attended information of the query fragment is encoded as the weighted sum of all key fragment features. The similarity between each query fragment and its attended information is thus aggregated as the similarity measurement between the query and the retrieval candidates.

In ideal cases, well-trained cross-modal attention models will attend to the semantically relevant key fragments by assigning large attention weights to them, and ignore irrelevant fragments by producing small attention weights. Take Figure~\ref{fig:fig_1}~(b) as an example: when ``dog" is used as a query fragment, the cross-modal attention model is supposed to output large attention weights for all image regions containing the dog, and small attention weights for other irrelevant image fragments. However, since the cross-modal attention models of most existing image-text matching methods are trained in a purely data-driven manner and do not receive any explicit supervision or constraints, the learned attention models may not be able to precisely attend to the relevant contents. As shown in Figure~\ref{fig:fig_1}~(a), the learned SCAN model~\cite{lee2018stacked}, a state-of-the-art cross-modal attention-based image-text matching model, fails to attend to the relevant image regions containing the dog's main body when using the word ``dog" as the query fragment. This example illustrates a false negative case, i.e., a low attention ``recall". Additionally, a learned cross-modal attention model might also suffer from false positives (low attention ``precision"). As shown in Figure~\ref{fig:fig_1}~(c), when using ``helmet" as the query fragment, the SCAN model assigns large attention weights to the irrelevant human body and background areas. A possible solution to these limitations is to rely on manually generated attention map ground truth to supervise the training process of cross-modal attention models~\cite{qiao2018exploring,zhang2019interpretable}. However, annotating attention distributions is an ill-defined task, and will be labor-intensive. 

% that are considered as contexts to the helmet. Admittedly, contexts provide clues that are critical to detecting small objects. However, they also include noise and interference in precisely capturing the desired contents

To this end, we propose two learning constraints, namely \textbf{Contrastive Content Re-sourcing (CCR)} and \textbf{Contrastive Content Swapping (CCS)}, to supervise the training process of cross-modal attentions. Figure~\ref{fig:fig_2} gives an overview of our method. CCR enforces a query fragment to be more relevant to its attended information than to the reversed attended information, which is generated by calculating the weighted sum of key fragments using reversed attention weights (details in Section~\ref{sec:method:ccr}). It can guide a cross-modal attention model to assign large attention weights to the relevant key fragments and small weights to irrelevant fragments. On the other hand, CCS further encourages a cross-modal attention model to ignore irrelevant key fragments by constraining the attended information to be more relevant to the corresponding query fragment than to a negative query fragment. In the example shown in Figure~\ref{fig:fig_2} (c), by using the word ``grass" as a negative query fragment, the attention weights assigned to regions containing grass will be diminished so that a more accurate attention map is generated. The proposed constraints are plug-in training strategies that can be easily integrated into existing cross-modal attention-based image-text matching models.

We evaluate the performance of the proposed constraints by incorporating them into four state-of-the-art cross-modal attention-based image-text matching networks~\cite{lee2018stacked,liu2019focus, wang2019position, diao2021similarity}. Additionally, in order to quantitatively compare and measure the quality of the learned attention models, we propose three new attention metrics, namely \textbf{Attention Precision}, \textbf{Attention Recall} and \textbf{Attention F1-Score}. The experimental results on both MS-COCO~\cite{lin2014microsoft} and Flickr30K~\cite{young2014image} demonstrate that these constraints significantly improve image-text matching performances and the quality of the learned attention models.% of these methods.

To sum up, the main contributions of this work include: (i) we propose two learning constraints to supervise the training of cross-modal attention models in a contrastive manner without requiring additional attention annotations. They are plug-in training strategies and can be easily applied to different cross-modal attention-based image-text methods; (ii) we introduce the attention metrics to quantitatively evaluate the quality of learned attention models, in terms of precision, recall, and F1-Score; (iii) we validate our approach by incorporating it into four state-of-the-art attention-based image-text matching models. Extensive experiments conducted on two publicly available datasets demonstrate its strong generality and effectiveness.

\section{Related Work}
\textbf{Image-Text Matching.}
The task of image-text matching is well-explored yet challenging. Its main challenge is how to measure the similarity between texts and images.  Early approaches propose to measure the similarity at the global level~\cite{kiros2014unifying,frome2013devise,zhang2018deep,faghri2017vse++}. Specifically, these methods first train an image encoder and a text encoder to embed the global information of images and sentences into feature vectors, and then measure the similarity between images and sentences by calculating the cosine  similarity between the corresponding feature vectors. For example, by using the triplet ranking loss with hard negative samples, Faghri \etal{}~\cite{faghri2017vse++} train a VGG-based image encoder~\cite{simonyan2014very} and a GRU-based text encoder~\cite{chung2014empirical}, respectively. %and then the similarity score between an image and a sentence was obtained by calculating the cosine similarity between their feature vectors, which were extracted by the trained encoders. 
One major limitation of these methods is that they failed to capture fine-grained image-text relevance. 

To address this limitation, recent studies propose to apply the cross-modal attention mechanism to measure the similarity between texts and images at the fragment level~\cite{liu2019focus,lee2018stacked,wang2019position,xu2020cross}. Typically, given an image and a sentence, these methods first extract embeddings on object regions from the image by feeding it into an object detection model, such as Faster R-CNN~\cite{ren2015faster}, and embed each word of the sentence by using recurrent neural networks. Then the relevant regions of each word and the relevant words of each region are inferred by leveraging the text-to-image and image-to-text attention, respectively. The similarity between each fragment (word or image region) and its relevant information is calculated and aggregated as the final similarity score between the image and sentence. Although these methods have achieved notable results, the learning process of these cross-modal attention models could be sub-optimal due to the lack of direct supervision, as discussed in Section~\ref{sec:intro}.

\textbf{Supervision on Learning Cross-Modal Attention.} The task of training cross-modal attention models with proper supervision has drawn growing interests. The main challenge lies in how to define and collect supervision signals. Qiao \etal{} \cite{qiao2018exploring} first train an attention map generator on a human annotated attention dataset and then apply the attention map predicted by the generator as weak annotations. Liu \etal{}~\cite{liu2017attention} leverage human annotated alignments between words and corresponding image regions as supervision. Similar to~\cite{liu2017attention}, image local region descriptions and object annotations in Visual Genome~\cite{krishna2017visual} are used for generating attention supervision~\cite{zhang2019interpretable}. These methods obtain attention supervision from different forms of human annotations, such as word-image correspondence and image local region annotations. By contrast, we provide attention supervision by constructing pair-wise samples in a contrastive learning manner which does not require additional manual attention annotations.

% The task of training cross-modal attention models with proper supervision has drawn growing interests. The main challenge lies in how to define and collect supervision signals. \citet{qiao2018exploring} first trains an attention map generator on a human annotated attention dataset and then applies the attention map predicted by the generator as weak annotations.~\citet{liu2017attention} leverages human annotated alignments between words and corresponding image regions as supervision. Similar to~\cite{liu2017attention}, image local region descriptions and object annotations in Visual Genome~\cite{krishna2017visual} are used for generating attention supervision~\cite{zhang2019interpretable}. These methods obtain attention supervision from different forms of human annotations, such as word-image correspondence and image local region annotations. By contrast, we provide attention supervision by constructing pair-wise samples in a contrastive learning manner which does not require additional manual attention annotations.

\section{Methodology}
\label{sec:method}
% The process of leveraging Contrastive Content Re-sourcing and Contrastive Content Swapping for contrastive learning is shown in Figure~\ref{fig:fig_2}. We formulate our task follows the setting of image-text matching since it is our main application, and similar process can be applied to other cross-modal vision-language tasks, as discussed in Section~\ref{sec:discussion}.

%In this section, our problem formulation follows the setting of a image-text matching task since it is our main application, and similar process can be applied to other cross-modal vision-language tasks as well.
%In this section, we instantiate our proposed constraints for the text-image matching task. We begin by introducing the general attention framework used in text-image matching in Section~\ref{method:gerneral_att}. We then present our proposed CCR and CCS in Sections~\ref{method:CCR} and~\ref{method:CCS}, respectively. At last but not least, our proposed attention metrics are introduced in Section~\ref{method:att_metric}. Note that the proposed method can be easily applied to other tasks that involve attention mechanisms by only minor adaptation. 

%-------------------------------------------------------------------------
\subsection{Cross-Modal Attention Model} %in Image-Text Matching}
\label{method:gerneral_att}
Given an image-sentence pair in image-text matching, they are first represented as fragments, i.e., individual  words and image regions. The fragments of the context modality are query fragments, and the fragments of the attended modality are key fragments. Each of these fragments is encoded as a vector. A cross-modal attention model takes these vectors as input, and infers the cross-modal relevance between each query fragment and all key fragments. The similarity score of the image-sentence pair is then calculated according to the obtained cross-modal relevance.

%we refer the attended modality as key modality, the other as query modality. In addition,
Let $q_i$ and $k_j$ refer to the feature representation of the $i$-th query and $j$-th key fragments, respectively. %For example, when attending image regions with respect to each word, $q_i$ and $k_j$ is the $i$-th word feature and the $j$-th image feature. 
The cross-modal attention model first calculates $k_j$'s attention weight with respect to $q_i$ as follows:
\begin{equation}
	\label{eq:att_weight}
    \begin{aligned}
		e_{i,j}&=f_{att}(q_{i},k_j),\\
        w_{i,j}&=\frac{\exp (e_{i,j})}{\sum_{j \in K}\exp (e_{i,j})},
    \end{aligned}
\end{equation}
where $f_{att}$ is the attention function whose output is a scalar $e_{i,j}$ that measures the cross-modal relevance between $q_{i}$ and $k_j$; $K$ is a set of indexes of all key fragments; $w_{i,j}$ is $k_j$'s attention weight with respect to $q_i$. 

$q_i$'s \textbf{attended information} (i.e., $q_i$'s relevant cross-modal information) is defined as the \textbf{attention feature} $a_i$ using the weighted sum of key fragment features in the following equation:
\begin{equation}
    \label{eq:att_feature}
        a_i =\sum_{j \in K} \left(w_{i,j}\cdot k_j \right).
\end{equation}

The similarity score between the image $I$ and the sentence $T$ is then defined as:
\begin{equation}
    \label{eq:sim_score}
        S(I, T) = AGG_{i \in Q}(Sim(q_{i}, a_{i})),
\end{equation}
where $Q$ denotes the set of indexes of all query fragments; $Sim$ is the similarity function; $AGG$ is a function that aggregates similarity scores among all query fragments, such as the average pooling function~\cite{lee2018stacked}.

The most widely used loss function for this task is the triplet ranking loss with hard negative sampling~\cite{faghri2017vse++} defined as:
\begin{multline}
    \label{eq:loss_rank}
    \ell_{rank} = [S(I, \hat{T}) - S(I, T) + \gamma_1 ]_+\\ 
        + [S(\hat{I}, T) - S(I, T) + \gamma_1 ]_+,
\end{multline}
where $\gamma_1$ controls the margin of similarity difference; the matched image $I$ and the sentence $T$ form a positive sample pair, while $\hat{T}$ and $\hat{I}$ represent the hardest negative sentence and image for the positive sample pair as defined by~\cite{faghri2017vse++}. $\ell_{rank}$ enforces the similarity between the anchor image $I$ and its matched sentence $T$ to be larger than the similarity between the anchor image and an unmatched sentence by a margin $\gamma_1$. Vice versa for the sentence $T$.

However, this loss function works at the similarity level and does not provide any supervision for connecting cross-modal contents at the attention level. In other words, learning cross-modal attentions is a pure data-driven approach and lacks supervision. As a result, the learned cross-modal attention model could be sub-optimal.

\subsection{Contrastive Content Re-sourcing}
\label{sec:method:ccr}
% A desired property of a well-learned attention model is that, given a query fragment, the query fragment is more relevant to its attended information than to its ignored information. The main idea of the proposed Contrastive Content Re-sourcing (CCR) is to explicitly constrain attention models to learn this property in a contrastive learning manner, as shown in Figure~\ref{fig:fig_2} (b). 

% A desired property of a well-learned attention model is that, for a query fragment, the attention model attend on information that is semantically relevant to the query fragment's, and ignored information that is irrelevant to the query fragment. In other words, for a query fragment, the semantic relevance between it and its attended information is larger than that between it and its ignored information. The main idea of the proposed Contrastive Content Re-sourcing (CCR) is to explicitly constrain attention models to learn this property in a contrastive learning manner. 

A desired property of a well-learned cross-modal attention model is that, for a query fragment, the attention model should assign large attention weights to the key fragments that are relevant to the query fragment, and assign small attention weights to the key fragments that are irrelevant to the query fragment. The Contrastive Content Re-sourcing (CCR) constrain is proposed to explicitly guide attention models to learn this property. It enforces a query fragment to be more relevant to its attended information than to its reversed attention information. For example, as shown in Figure~\ref{fig:fig_2} (b), the query ``dog" is required to be more relevant to its attended information than to the reversed attended information which contains the person and trees. 

To be specific, given a query fragment $q_i$, its attended information is embedded as the attention feature $a_i$. Its reversed attention information is encoded by the vector $\hat{a}_i$, which is obtained by reversing attention weights and calculating weighted sum of key fragment features based on the reversed attention weights, as shown in Equation~\ref{eq:att_ign}:
\begin{equation}
	\label{eq:att_ign}
    \begin{aligned}
        \hat{w}_{i,j} &= \frac{1 - w_{i,j}}{\sum_{j \in K}(1 - w_{i,j})},\\
        \hat{a}_{i} &= \sum_{j \in K} \left(\hat{w}_{i,j}\cdot k_j \right) ,
    \end{aligned}
\end{equation}
where $\hat{w}_{i,j}$ is the reversed attention weight of the key fragment $k_j$ with respect to the query fragment $q_i$.

We use the similarity function $Sim$ to measure the relevance between the query fragment and either the attention feature or reversed one. Therefore, the loss function for CCR is defined as:
% \begin{equation}
%  AGG_{i}^{i \in I} (R(q_{i}, b_{i})) > AGG_{i}^{i \in A}(R(q_{i}, c_{i}))
% \end{equation}
\begin{equation}
    \label{loss_CCR}
         \ell_{CCR} = [Sim(q_{i}, \hat{a}_{i}) - Sim(q_{i}, a_{i}) + \gamma_2 ]_+,
\end{equation}
where $\gamma_2$ controls the similarity difference margin.

Intuitively, in order to minimize this loss, a cross-modal attention model should assign large attention weights to relevant key fragments to increase $q_i$'s relevant information ratio in $a_i$ and decrease that contained in $\hat{a}_i$. The attention model will also learn to assign small attentions weights to irrelevant key fragments to diminish $q_i$'s irrelevant information ratio in $a_i$ and increase that in $\hat{a}_i$.

% Otherwise, $ign_i$ will contains $q_i$ relevant information, and $a_i$ will contains $q_i$ irrelevant information. As a result, the similarity difference will smaller than the margin $\gamma_2$ and the attention model will be punished by $\ell_{CCR}$.

% $l_{CCR}$ bridges the gap between the task's objectives (similarity score for image-text pairs) and intermediate learning process (attention weights). In other words, the model is encouraged to assign higher weights to the key fragments whose contents are more relevant to the query fragments, which enforces the motivation for applying attention models in the first place.

%We devise the bellow loss function to guide attention models to learn to satisfy the CAC:  

% \begin{equation}
%     \label{eq:s_value}
%          \ell_{CAC} = |AGG_{i}^{i \in Q}(R(q_{i}, c_{i})) - AGG_{i}^{i \in Q}(R(q_{i}, b_{i})) + \gamma_1 |
% \end{equation}
% where $\gamma_1 \geq 0$ is a variable controlling the margin between attention values.

\subsection{Contrastive Content Swapping}
\label{sec:method:ccs}

% The motivation behind the CCS constraints is that, in order to minimize $\ell_{CCS}$, attention models will learn to diminish the attention weights of the key fragments that are relevant to $\bar{q_{p}}$
As shown in Figure~\ref{fig:fig_1}~(c), attention models could assign large attention weights to both relevant and irrelevant key fragments. In such cases, the CCR constraint might not be able to fully address these false-positive scenarios because the query fragment can be more relevant to its attended information than to its reversed attention information. Therefore, we propose the Contrastive Content Swapping (CCS) constraint to address this problem. It constrains a query fragment's attended information to be more relevant to the query fragment than to a negative query fragment.

% Consider the case where a learned attention model assigns high attention weights to both relevant key fragments and some irrelevant key fragments with respect to a query fragment. Obviously, if the attention model is enforced to make relevance between the query fragment's attended information and the query fragment be large than that between the attended information and a fragment that correspondent to the irrelevant fragments with large attention weights, it will dismiss the attention weights for irrelevant fragments. It motivates the BB, 
% We will first introduced the case when attending to image region with respect to each word, and then introduced the case when attending to word with respect to each region.  
% The main idea of Contrastive Content Swapping is similar to triplet loss based metric learning, where the similarity between a positive data pair (anchor image and its caption) is enforced to be higher than the similarity between a negative data pair (anchor image and an unmatched caption). CCS applies the same training logic at the fragment-level.

% Specifically, we first generate negative samples, referred to \textbf{swapped query fragments}, by ``swapping'' the contents in the original query fragments with visually or semantically different contents from the same modality. 

Specifically, given a query fragment $q_i$, we first sample its negative query fragment $\hat{q}_i$ from a predefined set $\hat{Q}_i$ which contain all negative query fragments with respect to $q_i$. The relevance between the attended information and either the query fragment or the negative query fragment is also measured by the similarity function $Sim$. Then the CCS constraint's loss function $\ell_{CCS}$ is defined as:
\begin{equation}
    \label{eq:s_value_y}
         \ell_{CCS} =  [ Sim(\hat{q}_{i}, a_{i}) - Sim(q_{i}, a_{i})  + \gamma_{3} ]_+,
\end{equation}
where $\gamma_3$ is the margin parameter.

% $\mathcal{P}_i$ is the set of swapped query fragments for $q_i$; $\bar{q_{p}}$ is the embedding of the $p$-th fragment in $\mathcal{P}_i$.

The CCS constraint will enforce the cross-modal attention model to diminish the attention weights of the key fragments that are relevant to $\hat{q}_{i}$. As a result, the information that is relevant to $\hat{q}_i$ but irrelevant to $q_i$ is eliminated.

By incorporating the CCR and CCS constraints for image-text matching, we obtain the full objective function by Equation~\ref{eq:s_value_t}, where $\lambda_{CCR}$ and $\lambda_{CCS}$ are scalars that control the contributions of CCR and CCS, respectively:
\begin{equation}
    \label{eq:s_value_t}
         \ell = \ell_{rank} + \lambda_{CCR} \cdot \ell_{CCR} + \lambda_{CCS} \cdot \ell_{CCS}.
\end{equation}

% \begin{equation}
%  R(luq_{i}, a_{i}) > R(\bar{q}, a_{i}),
% \end{equation}
% where $\bar{q}$ is the feature representation of a word that is semantically different with $q_{i}$.

% Consider the case when the learned attention model assigns high attention weights to the key fragments that are relevant to $\bar{q}$ or $q_{i}$. As a result, $a_{i}$ is relevant to both $q_{i}$ and $\bar{q}$. In such case, the CCS will guide the attention models to decrease the attention weights to key fragments that relevant to $\bar{q}$, and increase the attention weights for key fragments that relevant to $q_i$. 

\subsection{Attention Metrics}
\label{method:att_metric}

Previous studies~\cite{lee2018stacked,liu2019focus} focus on qualitatively evaluating the attention models by visualizing attention maps. These approaches cannot serve as standard metrics for comparing attention correctness among different models. Therefore, we propose Attention Precision, Attention Recall and Attention F1-Score, to quantitatively evaluate the performance of learned attention models. Attention Precision is the fraction of attended key fragments that are relevant to the correspondent query fragment, and Attention Recall is the fraction of relevant key fragments that are attended. Attention F1-Score is a combination of the Attention Precision and Attention Recall that provides an overall way to measure the attention correctness of a model.

In this paper, we only evaluate the attention models that use texts as the query fragments. This is because text encoders used in the evaluated models~\cite{lee2018stacked, wang2019position,liu2019focus,diao2021similarity} are GRUs~\cite{chung2014empirical} or Transformers~\cite{vaswani2017attention}, where defining the relevant and irrelevant key text fragments of a query region fragment could be difficult since the text fragments will be updated to include global information by the text encoder.

Given a matched image-text pair, an image fragment $v$ is labeled as a relevant fragment of the text fragment $t$ if the Intersection over Union (IoU)\footnote{Given two bounding boxes, the IoU score between them is calculated as the ratio of their joint area to their union area.} between $v$ and the correspondent region\footnote{The correspondent regions of a word $t$ are the regions that contain the object described by $t$.} of $t$ is larger than a threshold $T_{IoU}$. In addition, $v$ is regraded as an attended fragment by $t$ if $v$'s attention weight with respect to $t$ is larger than a threshold $T_{Att}$. Let $A$ and $R$ be the sets of attended and relevant image fragments of $t$. $t$'s Attention Precision ($AP$), Attention Recall ($AR$), and Attention F1-Score ($AF$) are defined as:
% \begin{equation}
% \label{eq:att_metric}
% \begin{aligned}
% AP = \frac {|A \cap R|}{|A|}, \\
% AR = \frac {|A \cap R|}{|R|}, \\  
% AF = 2 \times \frac {AP \times AR}{AP + AR}. \\
% \end{aligned}
%\end{equation}

\begin{equation}
\label{eq:att_metric}
AP = \frac {|A \cap R|}{|A|}, AR = \frac {|A \cap R|}{|R|}, AF = 2 \times \frac {AP \times AR}{AP + AR}.
\end{equation}

% AP_j &= \frac{ \text{\# of the attended relevant regions}}{\text{\# of the attended regions}}, \\
% AR_j &= \frac{\text{\# of the attended relevant regions}}{\text{\# of the relevant regions}} ,\\

The annotations ~\cite{plummer2015flickr30k} that are used to calculate attention metrics provide the correspondence between noun phrases and image regions. A noun phrase might contain multiple words, and different words could correspond to the same image region. In order to obtain the overall attention metrics of a learned attention model, we first calculate the attention metrics at word-level, and use the maximal values within each phrase as the phrase-level metrics. The overall attention metrics are then obtained by averaging the phrase-level metrics.

% Specially, the recall of a attention model is defined as the ratio of the number of relevant key fragments that are assigned with large attention weights to the number of relevant key fragments, and the precision is defined as the the ratio of the number of relevant key fragments that assigned with large attention weights to the number of key fragments are assigned with large attention.  

%---------------- tables reformatting

% %---------------- tables reformatting

% \begin{table*}[!t]
% \centering
% \begin{tabular}{@{}llllllll@{}}
% \toprule
%                           & \multicolumn{3}{c}{Sentence Retrieval}                                       & \multicolumn{3}{c}{Image Retrieval}                                          &       \\
% \multicolumn{1}{c}{Method} & \multicolumn{1}{c}{R@1} & \multicolumn{1}{c}{R@5} & \multicolumn{1}{c}{R@10} & \multicolumn{1}{c}{R@1} & \multicolumn{1}{c}{R@5} & \multicolumn{1}{c}{R@10} & rsum  \\ \midrule
% SCAN  & 55.0 &	80.7 & 88.2	& 41.8	& 69.9	& 78.7	&414.3 \\
% SCAN + CCR + CCS  & \textbf{55.3} & \textbf{82.4}	& \textbf{90.2}	& \textbf{45.1}	& \textbf{71.5}	& \textbf{79.7}	& \textbf{424.2} \\
% \midrule

% BFAN & 57.5 &	84.7&	91.5&	45.0&	71.7&	80.0&	430.4 \\
% BFAN + CCR + CCS  & \textbf{59.9}&	\textbf{85.7}&	\textbf{92.0}&	\textbf{46.3}&	\textbf{72.6}&	\textbf{80.9}&	\textbf{435.2} \\ \bottomrule
% \end{tabular}
% \caption{Results of testing models trained on MS-COCO sentence retrieval and image retrieval tasks on the Flickr30K test set. R@K refers to Recall@K.}
% \label{tbl:domain_adpation}
% \end{table*}

\section{Experiments}
\label{sec:exp}
\subsection{Datasets and Evaluations}
\label{dataset_setting}

\textbf{Datasets.} We evaluate our method on two public image-text matching benchmarks: Flickr30K~\cite{young2014image} and MS-COCO~\cite{lin2014microsoft}. Flickr30K~\cite{young2014image} dataset contains 31K images, each of which is annotated with 5 captions. Following the setting of~\cite{liu2019focus,lee2018stacked}, we split the dataset into 29K training images, 1K validation images, and 1K testing images. The MS-COCO dataset used for image-text matching consists of 123,287 images, each of which includes 5 human-annotated descriptions. Following~\cite{liu2019focus,lee2018stacked}, the dataset is divided into 113,283 images for training, 5K images for validation, and 5K images for testing. 

\textbf{Evaluation Metrics.}
Following~\cite{liu2019focus,lee2018stacked,wang2019position}, we measure the performance of both \textbf{Image Retrieval} and \textbf{Sentence Retrieval} tasks by calculating recalls at different K values (R@K, K = 1, 5, 10), which are the proportions of the queries whose top-K retrieved items contain their matched items. We also report \emph{rsum}, which is the summation of all R@K values for a model. On the Flickr30K dataset, we report results on the 1K testing images. On the MS-COCO dataset, we report results through averaging over 5-folds 1K test images (referred to MS-COCO  1K), and testing on the full 5K test images (referred to MS-COCO  5K) following the standard evaluation protocol~\cite{lee2018stacked, liu2019focus,wang2019position}. 

To compute the attention metrics, $T_{IoU}$ is set as 0.4, and the results for other values of $T_{IoU}$ can be found in the supplementary material. The possible values of $T_{Att}$ are uniformly chosen between 0 and 0.1 with the interval of 0.01. We set the range of $T_{Att}$ based on the experimental results that when achieving the best Attention F1-Score the $T_{Att}$is ranging from 0 to 0.1. We calculate the Attention Precision, Attention Recall and Attention F1-Score for each value of $T_{Att}$, and then report the precision-recall (PR) curves and the best Attention F1-Score with its correspondent Attention Precision and Attention Recall. 

% For the settings with our proposed constraints, models whose constraint margin ($\gamma_{2}$ or $\gamma_{3}$) and loss weight factor ($\lambda_{CCR}$ or $\lambda_{CCS}$) are trained and report best performance

\begin{table}[t]
\centering
\resizebox{\linewidth}{!}{
\begin{tabular}{@{}llllllll@{}}
\toprule
                           & \multicolumn{3}{c}{Sentence Retrieval}                                       & \multicolumn{3}{c}{Image Retrieval}                                          &       \\
\multicolumn{1}{c}{Method} & \multicolumn{1}{c}{R@1} & \multicolumn{1}{c}{R@5} & \multicolumn{1}{c}{R@10} & \multicolumn{1}{c}{R@1} & \multicolumn{1}{c}{R@5} & \multicolumn{1}{c}{R@10} & rsum  \\ \midrule

SCAN~\cite{lee2018stacked}  & 67.2 & 90.7 & 94.8 & 48.4  & 77.6 & 84.9 & 463.6 \\
+ CCR & 67.8 & 91.1 & 95.0 & 49.4 & 77.6  & 85.3 & 466.2 \\
+ CCS  & \textbf{69.1}  & 91.1	& \textbf{95.4}	& 50.8	& 78.4	& 85.6	& 470.4 \\
+ CCR \& CCS  & 68.8	& \textbf{91.6}	& 95.3	& \textbf{51.1}	& \textbf{79.0}	& \textbf{86.5}	& \textbf{472.3} \\
\midrule

PFAN~\cite{wang2019position}  & 69.7 & 90.2 & 94.1 & 50.1  & 78.6 & 86.0 & 468.7 \\
+ CCR & 70.3 & 90.5 & 94.7 & 51.9 & 79.4  & 86.7 & 473.5 \\
+ CCS  & 70.3  & 90.9	& 95.2	& 51.9	& 79.2	& 86.5	& 474.0 \\
+ CCR \& CCS  & \textbf{70.9} & \textbf{91.8} & \textbf{95.6}	& \textbf{52.5}	& \textbf{79.6}	& \textbf{86.9}	& \textbf{477.3} \\
\midrule

BFAN~\cite{liu2019focus} & 70.7	& 92.3	& \textbf{96.3}	& 51.8	& 79.3	& 85.9	& 476.3 \\

+ CCR & 71.7 & 92.8	& 96.0	& \textbf{53.2}	& \textbf{80.5}	& \textbf{87.1}	& 481.3 \\

+ CCS  & 71.0 & 93.2 & 96.0	& 52.6	& 79.4	& 86.4	& 478.6 \\

+ CCR \& CCS  & \textbf{72.0} &\textbf{93.4}	& 96.2	&53.1	&80.3	&86.9	&\textbf{481.9} \\
\midrule

SGRAF~\cite{diao2021similarity}  & 77.8  & 94.5 & 96.8 & 59.0 & 82.9 & 88.6 & 499.6 \\
+ CCR & 78.0 & \textbf{95.2}  & 97.2 & 59.5 & 83.1 & 88.7 & 501.7 \\
+ CCS  & 78.3	& 94.6	& 97.4 & 59.6  & 83.5	& \textbf{89.0} & 502.4\\
+ CCR \& CCS  & \textbf{79.3}	& \textbf{95.2}	& \textbf{98.0} & \textbf{59.8} & \textbf{83.6} & 88.8 & \textbf{504.7}\\
\bottomrule

\end{tabular}}

\caption{Results of the sentence retrieval and image retrieval tasks on the Flickr30K test set.}
\label{tbl:f30k}
\end{table}

\begin{table}[t]
\centering
\resizebox{\linewidth}{!}{
\begin{tabular}{@{}llllllll@{}}
\toprule
                          & \multicolumn{3}{c}{Sentence Retrieval}                                       & \multicolumn{3}{c}{Image Retrieval}                                          &       \\
\multicolumn{1}{c}{Method} & \multicolumn{1}{c}{R@1} & \multicolumn{1}{c}{R@5} & \multicolumn{1}{c}{R@10} & \multicolumn{1}{c}{R@1} & \multicolumn{1}{c}{R@5} & \multicolumn{1}{c}{R@10} & rsum  \\ \midrule

1K Test Images \\ \midrule

SCAN~\cite{lee2018stacked}  & 70.6	& 93.8	& \textbf{97.7}	& 54.1	& 86.0	& 93.4	& 495.6 \\

+ CCR   & 71.4	& \textbf{94.2}	& \textbf{97.7}	& 55.6	& 86.7	& 93.8	& 499.4 \\
+ CCS   & 71.1	& 94.0	& \textbf{97.7}	& \textbf{56.6}	& 87.2	& \textbf{94.0}	& 500.6 \\

+ CCR \& CCS  & \textbf{71.6}	& 94.0	& \textbf{97.7}	& 56.4	& \textbf{87.3}	& \textbf{94.0}	& \textbf{501.0} \\
\midrule

PFAN*~\cite{wang2019position}  & 74.5 & 95.4	& \textbf{98.6} & 59.8	& 88.8 & 94.8	& 511.9 \\

+ CCR*   & 74.4	& 95.3	& 98.3	& 60.5	& \textbf{89.1}	& 94.8	& 512.4 \\

+ CCS*   & 74.9	& \textbf{95.8}	& 98.3	& 60.8	& \textbf{89.1}	& 94.5	& 513.4 \\

+ CCR \& CCS*  & \textbf{75.2} & 95.6	& 98.2	& \textbf{61.2}	& 88.9	& \textbf{94.7}	& \textbf{513.8} \\
\midrule

BFAN~\cite{liu2019focus} &75.0	&95.0 	&98.2	&58.8 	&88.3	&94.4  &509.7\\

+ CCR &\textbf{75.2}	&95.3 &\textbf{98.3}	&60.1	&88.7	&\textbf{94.7}	&512.3 \\

+ CCS  &75.1	&95.3	&\textbf{98.3}	&59.6 	&88.5	&94.6	&511.4 \\

+ CCR \& CCS  &\textbf{75.2} &\textbf{95.5} 	&98.1	&\textbf{60.3}	&\textbf{88.8} &\textbf{94.7} &\textbf{512.6}\\ \midrule

SGRAF~\cite{diao2021similarity} &79.7 	&96.5 	&98.5 &63.3 & 90.1	&95.7 &523.8 \\

+ CCR &79.7	&\textbf{96.8}	&98.7 &63.8	&90.4 	&\textbf{95.9} 	&525.3 \\

+ CCS  &79.7	&\textbf{96.8} 	&\textbf{98.8} &63.8	&90.3 	&95.7 	&525.1 \\

+ CCR \& CCS  &\textbf{80.2}	&\textbf{96.8}	&98.7 &\textbf{64.3}	&\textbf{90.6} &95.8	&\textbf{526.4} \\ \midrule

5K Test Images \\ \midrule

SCAN~\cite{lee2018stacked}  & 47.2	& 77.6	& 87.7	& 34.7	& 65.2	& 77.3	& 389.7 \\

+ CCR   & 47.7	& 78.3	& \textbf{88.2}	& 36.2	& 66.6	& 78.2	& 395.2 \\
+ CCS   & 46.5	& \textbf{78.5}	& 88.0	& 36.5	& 66.6	& 78.3	& 394.4 \\

+ CCR \& CCS  & \textbf{47.9}	& 78.1	& \textbf{88.2}	& \textbf{36.9}	& \textbf{66.9}	& \textbf{78.4}	& \textbf{396.4} \\
\midrule

BFAN~\cite{liu2019focus} & 52.5	&80.3	&89.5	&37.5	&66.7	&78.1	&404.6 \\

+ CCR  &52.0	&\textbf{81.5} 	&89.9	&\textbf{38.7}	&\textbf{67.8}	&\textbf{78.8}	& 408.7 \\

+ CCS  &\textbf{53.8} 	&81.1	&89.9 	&38.0 	&67.3 	&78.5	&408.6 \\

+ CCR \& CCS  &53.4	&81.3 	&\textbf{90.1}	&38.4	&67.6	&78.6	&\textbf{409.4} \\
\midrule

SGRAF~\cite{diao2021similarity} &58.3 	&84.8 	&91.9 &41.8 &70.9 	&81.2  & 428.9  \\

+ CCR &59.2	&84.8	&92.0 &42.2	&71.1 	&81.7 	&431.0    \\

+ CCS  &58.6	&\textbf{85.0} 	&\textbf{92.2} &42.2	&71.2 	&81.6 	&430.8 \\

+ CCR \& CCS  &\textbf{59.7}	&\textbf{85.0}	&92.0 &\textbf{42.3}	&\textbf{71.4} &\textbf{81.9}	&\textbf{432.3} \\
\bottomrule

\end{tabular}}
\caption{Results of the sentence retrieval and image retrieval tasks on the MS-COCO test set. *Note that since the official implementation of PFAN only provides 1K images for testing, PFAN is tested \textbf{without} 5-fold cross-validation under the setting of 1K test images, and cannot be tested under the setting of 5K test images.}
\label{tbl:coco}
\end{table}

\subsection{Baselines and Implementation Details}
We evaluate the proposed constraints by incorporating them into the following state-of-the-art attention-based image-text matching models:

\begin{itemize}
\item \textbf{SCAN}~\cite{lee2018stacked} is a stacked cross-modal attention model to infer the relevance between words and regions and calculate image-text similarity.
\item \textbf{PFAN}~\cite{wang2019position} improves cross-modal attention models by integrating image region position information into them.
\item \textbf{BFAN}~\cite{liu2019focus} is a bidirectional cross-modality attention model which allows to attend to relevant fragments
and also diverts all the attention into these relevant fragments to
concentrate on them. %which removes irrelevant fragments from attended information based attention weights. 
\item \textbf{SGRAF}~\cite{diao2021similarity} first learns the global and local alignments between fragments by using cross-modal attention models, and then applies the graph convolutional networks~\cite{kipf2016semi} to infer relation-aware similarities based on the local and global alignments.

\end{itemize}

We apply the proposed constraints to one randomly sampled query fragment for each matched image-text pair, in order to reduce the computational cost. For a query word fragment, its negative query set $Q_i$ is consisted of the other words of its correspondent sentence. For a query region fragment, its $Q_i$ is set as the other regions of its correspondent image. The constraint loss weight factors $\lambda_{CCR}$ and $\lambda_{CCS}$ could be 0.1 or 1, and constraint similarity margins $\gamma_{2}$ and $\gamma_{3}$ are set to 0, 0.1 or 0.2. We train models with all possible combinations with constraint loss weight factors and similarity margins, and report the best results.

The experiments on Flickr30K and MS-COCO are conducted on the RTX8000 and A100 GPU, respectively. All the baselines are trained by their officially released codes. \footnote{https://github.com/kuanghuei/SCAN}~\footnote{https://github.com/CrossmodalGroup/BFAN}~\footnote{https://github.com/HaoYang0123/Position-Focused-Attention-Network}~\footnote{https://github.com/Paranioar/SGRAF}. All models are trained from scratch by completely following their original hyper-parameters settings such as the learning rate, batch size, model structure, and optimizer~\cite{lee2018stacked,liu2019focus, wang2019position,diao2021similarity}. More implementation details can be found in the supplementary materials.

\subsection{Experiments on Image-Text Matching}
\label{sec:exp:retrieval}
We start by evaluating the proposed approach for image and sentence retrieval tasks on both Flickr30K and MS-COCO datasets. Table~\ref{tbl:f30k} and Table~\ref{tbl:coco} show the results on the Flickr30K and MS-COCO datasets, respectively. We find that when the proposed CCR and CCS constraints are employed separately, they both achieve consistent performance improvements on all baselines and tasks. More importantly, all models achieve the best overall improvements (\emph{rsum}) when we apply both constraints. These results demonstrate the strong generality of our proposed constraints 
for different models and datasets. We also note that using CCR or CCS alone achieve better results than using both CCR and CCS under some metrics. One possible reason is that the CCR is expected to assign large attention weights to the key fragments that contain both irrelevant and relevant information. For example, it will attend to regions containing background and described objects. CCS tends to ignore these key fragments to decrease the attention weights on irrelevant information. As a result, using CCR and CCS together might result in conflicts in some rare cases, and using CCR (or CCS) alone may achieve slightly better results under some metrics.

\begin{table}[t]
\centering
\resizebox{0.8\linewidth}{!}{
\begin{tabular}{@{}lccc@{}}
\toprule
Method   & \makecell{Attention\\Precision} & \makecell{Attention\\Recall} & \makecell{Attention\\F1-Score} \\ \midrule
SCAN~\cite{lee2018stacked} & 32.79 & 65.30 & 39.96
           \\
+ CCR      & 36.30  & \textbf{66.80} &43.10            \\
+ CCS      & 37.28  & 64.97 & 43.38           \\
+ CCR \& CCS & \textbf{38.81} & 64.62  & \textbf{44.44}            \\ \midrule
BFAN~\cite{liu2019focus}           &46.08  &63.32    &48.91         \\
+ CCR      &50.21  &\textbf{64.20}    &\textbf{51.78}         \\
+ CCS      &49.16   &61.44      &49.74         \\
+ CCR \& CCS &\textbf{51.13}  &62.97  &51.73            \\ \midrule

SGRAF~\cite{diao2021similarity}   &44.54  &61.98    &47.91         \\
+ CCR      &45.22  &\textbf{64.07}    &49.12         \\
+ CCS      &47.43   &60.41    &49.20         \\
+ CCR \& CCS &\textbf{49.48}  &62.12  & \textbf{50.90}           \\ \bottomrule

\end{tabular}}
\caption{Results of Attention Precision, Attention Recall and Attention F1-Score (\%) of the SCAN, BFAN, and SGRAF models trained on the Flickr30K dataset. }
\label{tbl:att_score}
\end{table}

\begin{figure}[t]
 
  \subfloat[PR curves of SCAN]{%
    \includegraphics[clip,width=0.5\linewidth]{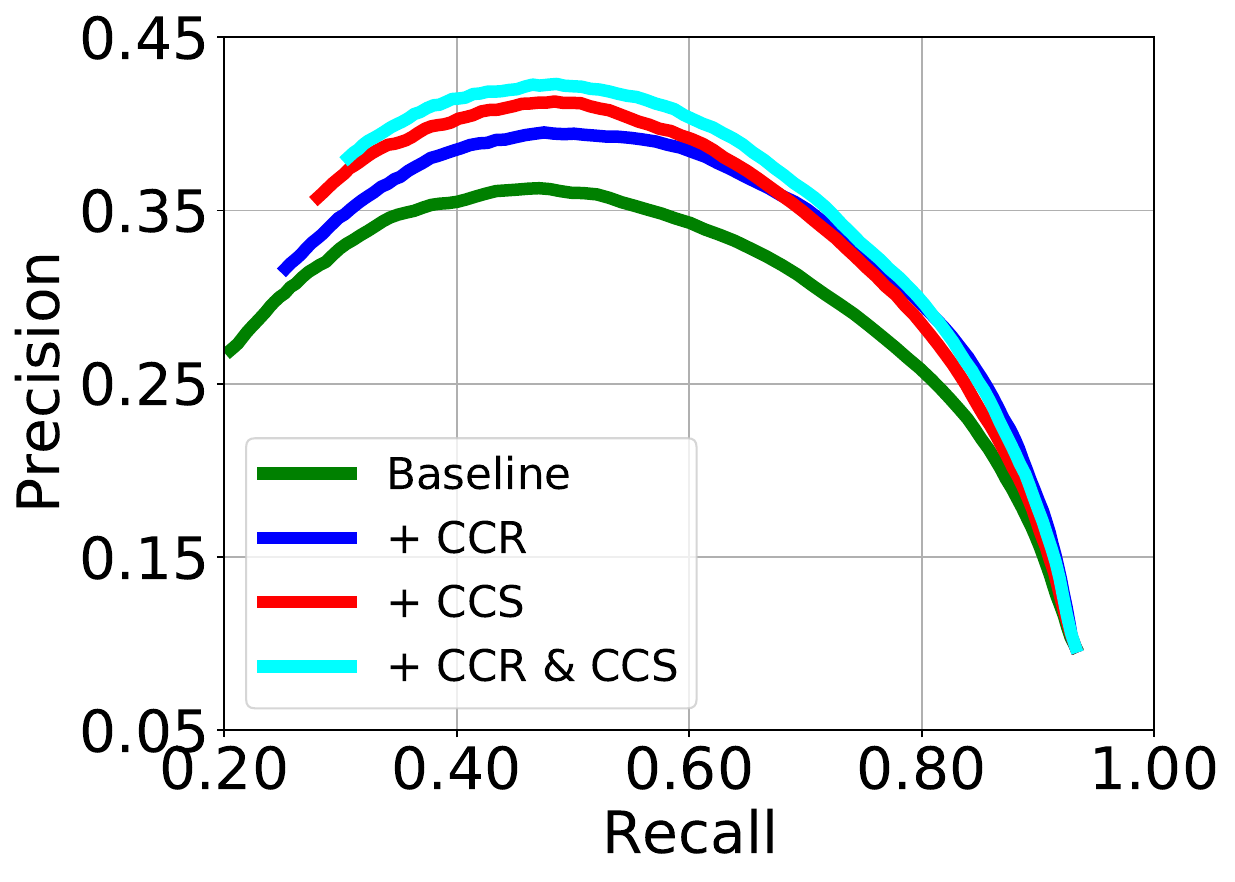}%
  }
  \subfloat[PR curves of BFAN]{%
    \includegraphics[clip,width=0.5\linewidth]{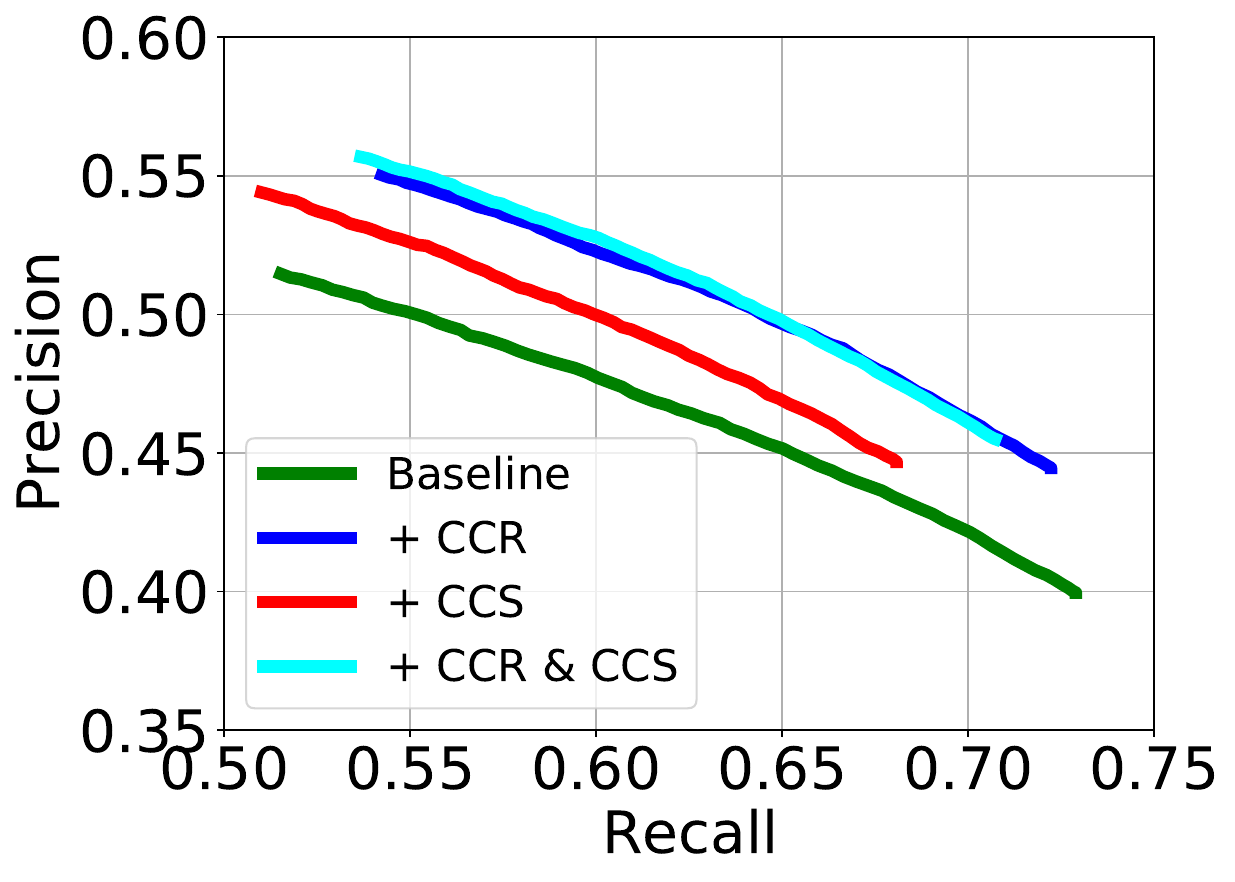}%
  }
  
  \centering
  \subfloat[PR curves of SGRAF]{%
    \includegraphics[clip,width=0.5\linewidth]{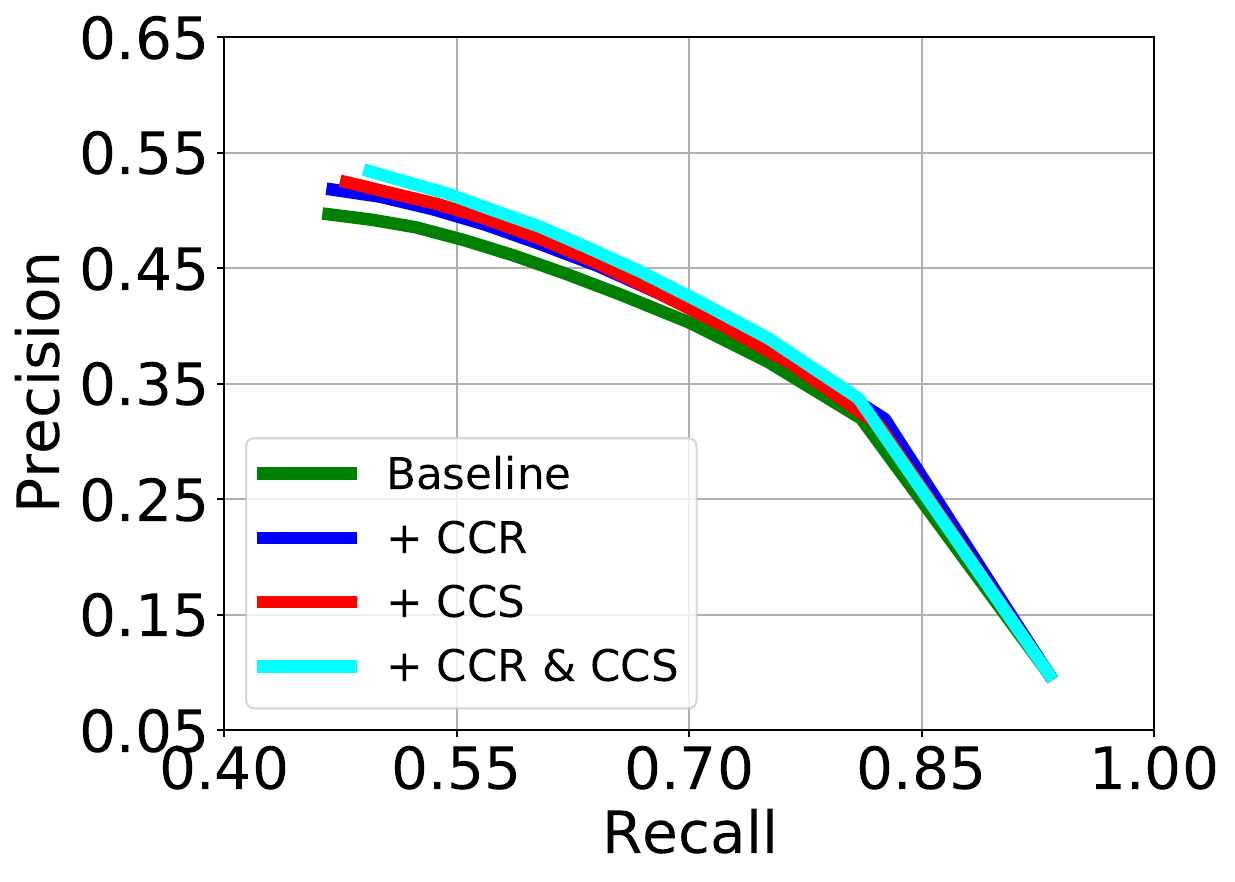}%
  }
  \caption{The Attention PR curves of the SCAN, BFAN, and SGRAF models trained on the Flickr30K dataset.}
  \label{fig:att_PR}
\end{figure}

\begin{figure*}[t]
\includegraphics[width=\textwidth]{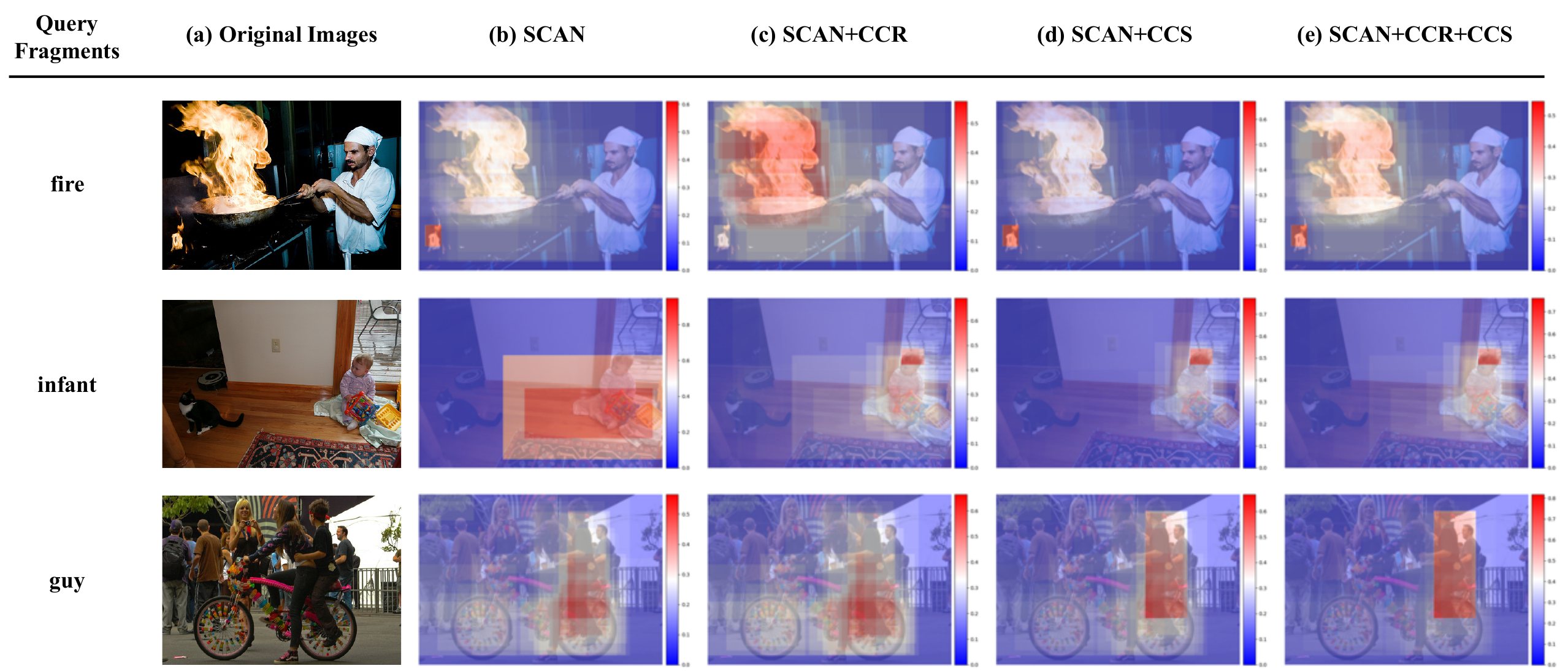}
\caption{Examples illustrating attended image regions with respect to the given words for the SCAN model on the Flickr30K dataset.}
\label{fig:att_vis_scan}
\end{figure*}

\begin{figure*}[t]
\includegraphics[width=\textwidth]{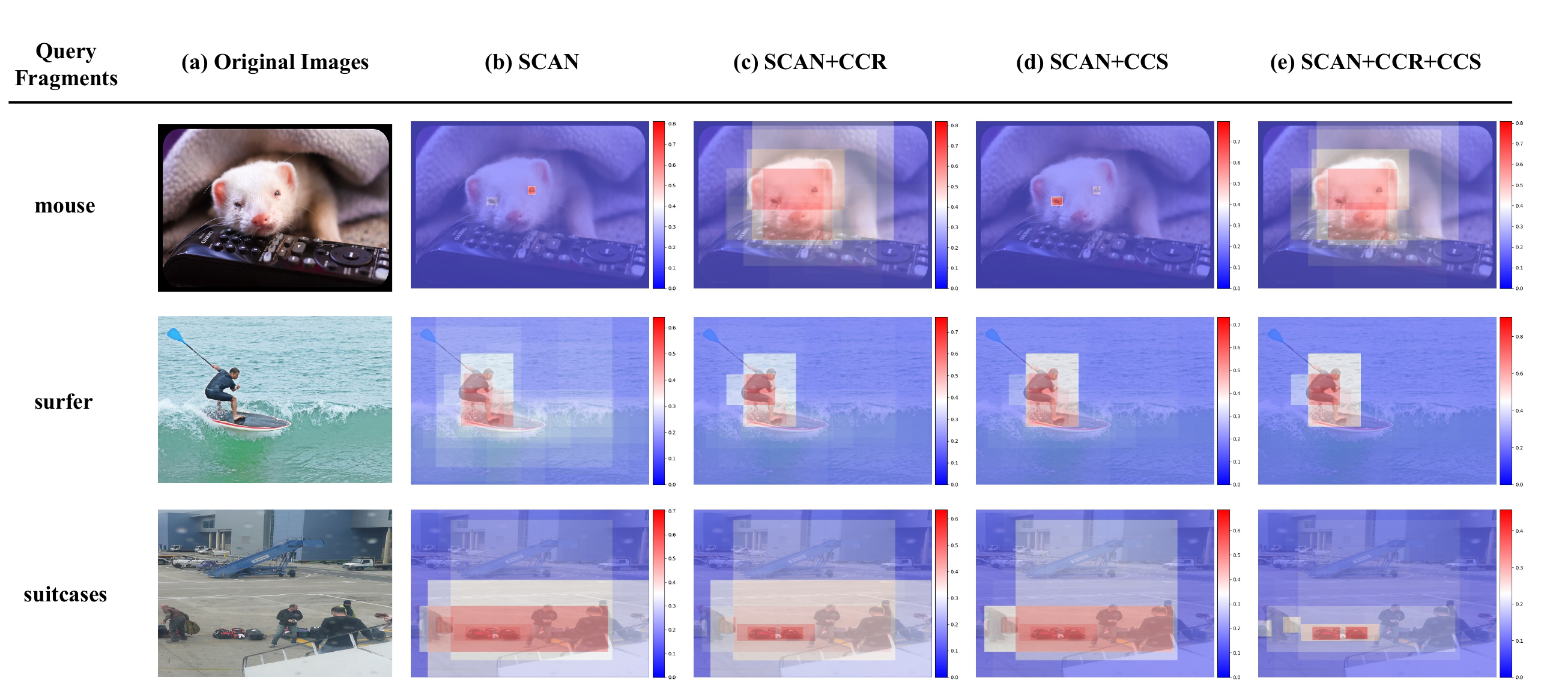}
\caption{Examples illustrating attended image regions with respect to the given words for the SCAN model on the MS-COCO dataset.}
\label{fig:att_vis_scan_coco}
\end{figure*}

\subsection{Attention Evaluation}
\textbf{Quantitative Analysis.} %We quantitatively evaluate the attention models using the proposed attention metrics, as shown in Table~\ref{tbl:att_score}. 
% In order to calculate the proposed attention metrics, cross-modal annotations that specify words or phrases and their image regions are required. 
We report the results on Flickr30K since it has publicly available cross-modal correspondence annotations~\cite{plummer2015flickr30k} while MS-COCO does not. We note that the results of PFAN are not reported because we cannot obtain the bounding boxes of the input image regions that are correspondent to the testing data provided by its official implementation. 

The attention metrics of SCAN, BFAN, and SGRAF are shown in Table~\ref{tbl:att_score}. We can see that applying CCR and CCS individually yields higher Attention F1-Score than both baseline methods, and this is consistent to the observations in Section~\ref{sec:exp:retrieval}. More interestingly, we can find that using CCR alone improves both Attention Precision and Attention Recall; using CCS alone mainly improves Attention Precision; combining both constraints further improves Attention Precision. These results show the constraints work as intended. Note that the slight decrease in Attention Recall caused by CCS might be due to the fact that CCS enforces attention models to ignore the regions containing both foreground objects and noise background. We also present the PR curves of SCAN, BFAN, and SGRAF in Figure~\ref{fig:att_PR} to demonstrate the impact of different $T_{Att}$ on Attention Precision and Attention Recall. We can observe that applying the proposed constraints yields consistently  better results than both baseline methods for different $T_{Att}$.

We further evaluate the relation between the image-text matching performance and the quality of learned attention models by calculating the Pearson correlation coefficient between Attention F1-Score and \emph{rsum} for each model. The obtained correlation coefficients of the SCAN, BFAN, and SGRAF models are 0.967, 0.992, and 0.941, respectively. The p-values are all less than 0.05. The results show that the image-text matching performance has strong positive correlation with the quality of learned attention models, which further demonstrate our motivation to propose the constraints.
\textbf{Qualitative Analysis.} We visualize the attention weights with respect to three sampled query word fragments on the Flickr30K and MS-COCO dataset. The results are shown in Figure~\ref{fig:att_vis_scan} and Figure~\ref{fig:att_vis_scan_coco}, respectively. More examples are provided in the supplementary material due to the space limitation. In the examples of the query word fragment ``fire" and ``mouse", the learned attention model of SCAN (see Column~(b)) fails to assign large attention weights to the most regions containing fire or mouse. By contrast, the CCR constraint (see Column~(c)) mitigates this issue by significantly increasing the attention weights assigned to the regions containing fire or mouse. The CCS constraint (see Column (d)) 
is less effective in these cases. In the cases of the query word fragment ``infant" and ``surfer", the learned attention model of SCAN (see Column~(b)) assigns large attention weights to both the irrelevant and relevant regions. In this case, the CCR constraint (see Column~(c)) cannot fully diminish the attention weights assigned to the regions irrelevant to ``infant'' and ``surfer'. In contrast, as shown in Column~(d), the attention weights assigned to irrelevant regions are largely diminished by the CCS constraint. In the examples of the query word ``guy'' and 'suitcases', they show that combining both constraints decreases the attention weights of the background regions (e.g., the surrounding areas of the ``guy'') more significantly than applying the them separately.

\section{Conclusions}
To tackle the issue of missing direct supervisions in learning cross-modal attention models for image-text matching, we introduce the constraints of CCR and CCS to supervise the learning of  attention models in a contrastive manner without requiring additional attention annotations. Both constraints are generic learning strategies that can be generally integrated into attention models. Furthermore, in order to quantitatively measure the attention correctness, we propose three new attention metrics. The extensive experiments demonstrate that the proposed constraints manage to improve the cross-modal retrieval performance as well as the attention correctness when integrated into four state-of-the-art attention models. For future work, we will explore on how to extend the proposed constraints to other cross-modal attention models based tasks, such as Visual Question Answering (VQA) and Image Captioning.

{\small
\bibliographystyle{ieee_fullname}
\bibliography{egbib}

\begin{thebibliography}{10}\itemsep=-1pt

\bibitem{chen2020adaptive}
Tianlang Chen, Jiajun Deng, and Jiebo Luo.
\newblock Adaptive offline quintuplet loss for image-text matching.
\newblock {\em arXiv preprint arXiv:2003.03669}, 2020.

\bibitem{chen2019uniter}
Yen-Chun Chen, Linjie Li, Licheng Yu, Ahmed El~Kholy, Faisal Ahmed, Zhe Gan, Yu
  Cheng, and Jingjing Liu.
\newblock Uniter: Learning universal image-text representations.
\newblock 2019.

\bibitem{chung2014empirical}
Junyoung Chung, Caglar Gulcehre, KyungHyun Cho, and Yoshua Bengio.
\newblock Empirical evaluation of gated recurrent neural networks on sequence
  modeling.
\newblock {\em arXiv preprint arXiv:1412.3555}, 2014.

\bibitem{diao2021similarity}
Haiwen Diao, Ying Zhang, Lin Ma, and Huchuan Lu.
\newblock Similarity reasoning and filtration for image-text matching.
\newblock In {\em Proceedings of the AAAI Conference on Artificial
  Intelligence}, volume~35, pages 1218--1226, 2021.

\bibitem{faghri2017vse++}
Fartash Faghri, David~J Fleet, Jamie~Ryan Kiros, and Sanja Fidler.
\newblock Vse++: Improving visual-semantic embeddings with hard negatives.
\newblock {\em arXiv preprint arXiv:1707.05612}, 2017.

\bibitem{frome2013devise}
Andrea Frome, Greg~S Corrado, Jon Shlens, Samy Bengio, Jeff Dean, Marc'Aurelio
  Ranzato, and Tomas Mikolov.
\newblock Devise: A deep visual-semantic embedding model.
\newblock In {\em Advances in neural information processing systems}, pages
  2121--2129, 2013.

\bibitem{huang2018bi}
Feiran Huang, Xiaoming Zhang, Zhonghua Zhao, and Zhoujun Li.
\newblock Bi-directional spatial-semantic attention networks for image-text
  matching.
\newblock {\em IEEE Transactions on Image Processing}, 28(4):2008--2020, 2018.

\bibitem{huang2017instance}
Yan Huang, Wei Wang, and Liang Wang.
\newblock Instance-aware image and sentence matching with selective multimodal
  lstm.
\newblock In {\em Proceedings of the IEEE Conference on Computer Vision and
  Pattern Recognition}, pages 2310--2318, 2017.

\bibitem{kipf2016semi}
Thomas~N Kipf and Max Welling.
\newblock Semi-supervised classification with graph convolutional networks.
\newblock {\em arXiv preprint arXiv:1609.02907}, 2016.

\bibitem{kiros2014unifying}
Ryan Kiros, Ruslan Salakhutdinov, and Richard~S Zemel.
\newblock Unifying visual-semantic embeddings with multimodal neural language
  models.
\newblock {\em arXiv preprint arXiv:1411.2539}, 2014.

\bibitem{krishna2017visual}
Ranjay Krishna, Yuke Zhu, Oliver Groth, Justin Johnson, Kenji Hata, Joshua
  Kravitz, Stephanie Chen, Yannis Kalantidis, Li-Jia Li, David~A Shamma, et~al.
\newblock Visual genome: Connecting language and vision using crowdsourced
  dense image annotations.
\newblock {\em International journal of computer vision}, 123(1):32--73, 2017.

\bibitem{lee2018stacked}
Kuang-Huei Lee, Xi Chen, Gang Hua, Houdong Hu, and Xiaodong He.
\newblock Stacked cross attention for image-text matching.
\newblock In {\em Proceedings of the European Conference on Computer Vision
  (ECCV)}, pages 201--216, 2018.

\bibitem{li2020oscar}
Xiujun Li, Xi Yin, Chunyuan Li, Pengchuan Zhang, Xiaowei Hu, Lei Zhang, Lijuan
  Wang, Houdong Hu, Li Dong, Furu Wei, et~al.
\newblock Oscar: Object-semantics aligned pre-training for vision-language
  tasks.
\newblock In {\em European Conference on Computer Vision}, pages 121--137.
  Springer, 2020.

\bibitem{lin2014microsoft}
Tsung-Yi Lin, Michael Maire, Serge Belongie, James Hays, Pietro Perona, Deva
  Ramanan, Piotr Doll{\'a}r, and C~Lawrence Zitnick.
\newblock Microsoft coco: Common objects in context.
\newblock In {\em European conference on computer vision}, pages 740--755.
  Springer, 2014.

\bibitem{liu2017attention}
Chenxi Liu, Junhua Mao, Fei Sha, and Alan Yuille.
\newblock Attention correctness in neural image captioning.
\newblock In {\em Proceedings of the AAAI Conference on Artificial
  Intelligence}, volume~31, 2017.

\bibitem{liu2019focus}
Chunxiao Liu, Zhendong Mao, An-An Liu, Tianzhu Zhang, Bin Wang, and Yongdong
  Zhang.
\newblock Focus your attention: A bidirectional focal attention network for
  image-text matching.
\newblock In {\em Proceedings of the 27th ACM International Conference on
  Multimedia}, pages 3--11, 2019.

\bibitem{nam2017dual}
Hyeonseob Nam, Jung-Woo Ha, and Jeonghee Kim.
\newblock Dual attention networks for multimodal reasoning and matching.
\newblock In {\em Proceedings of the IEEE Conference on Computer Vision and
  Pattern Recognition}, pages 299--307, 2017.

\bibitem{plummer2015flickr30k}
Bryan~A Plummer, Liwei Wang, Chris~M Cervantes, Juan~C Caicedo, Julia
  Hockenmaier, and Svetlana Lazebnik.
\newblock Flickr30k entities: Collecting region-to-phrase correspondences for
  richer image-to-sentence models.
\newblock In {\em Proceedings of the IEEE international conference on computer
  vision}, pages 2641--2649, 2015.

\bibitem{qiao2018exploring}
Tingting Qiao, Jianfeng Dong, and Duanqing Xu.
\newblock Exploring human-like attention supervision in visual question
  answering.
\newblock In {\em Proceedings of the AAAI Conference on Artificial
  Intelligence}, volume~32, 2018.

\bibitem{ren2015faster}
Shaoqing Ren, Kaiming He, Ross Girshick, and Jian Sun.
\newblock Faster r-cnn: Towards real-time object detection with region proposal
  networks.
\newblock {\em Advances in neural information processing systems}, 28:91--99,
  2015.

\bibitem{simonyan2014very}
Karen Simonyan and Andrew Zisserman.
\newblock Very deep convolutional networks for large-scale image recognition.
\newblock {\em arXiv preprint arXiv:1409.1556}, 2014.

\bibitem{vaswani2017attention}
Ashish Vaswani, Noam Shazeer, Niki Parmar, Jakob Uszkoreit, Llion Jones,
  Aidan~N Gomez, {\L}ukasz Kaiser, and Illia Polosukhin.
\newblock Attention is all you need.
\newblock In {\em Advances in neural information processing systems}, pages
  5998--6008, 2017.

\bibitem{wang2019position}
Yaxiong Wang, Hao Yang, Xueming Qian, Lin Ma, Jing Lu, Biao Li, and Xin Fan.
\newblock Position focused attention network for image-text matching.
\newblock {\em arXiv preprint arXiv:1907.09748}, 2019.

\bibitem{xu2020cross}
Xing Xu, Tan Wang, Yang Yang, Lin Zuo, Fumin Shen, and Heng~Tao Shen.
\newblock Cross-modal attention with semantic consistence for image-text
  matching.
\newblock {\em IEEE Transactions on Neural Networks and Learning Systems},
  2020.

\bibitem{young2014image}
Peter Young, Alice Lai, Micah Hodosh, and Julia Hockenmaier.
\newblock From image descriptions to visual denotations: New similarity metrics
  for semantic inference over event descriptions.
\newblock {\em Transactions of the Association for Computational Linguistics},
  2:67--78, 2014.

\bibitem{zhang2018deep}
Ying Zhang and Huchuan Lu.
\newblock Deep cross-modal projection learning for image-text matching.
\newblock In {\em Proceedings of the European Conference on Computer Vision
  (ECCV)}, pages 686--701, 2018.

\bibitem{zhang2019interpretable}
Yundong Zhang, Juan~Carlos Niebles, and Alvaro Soto.
\newblock Interpretable visual question answering by visual grounding from
  attention supervision mining.
\newblock In {\em 2019 IEEE Winter Conference on Applications of Computer
  Vision (WACV)}, pages 349--357. IEEE, 2019.

\end{thebibliography}
}

\clearpage

\includepdf[pages=1-]{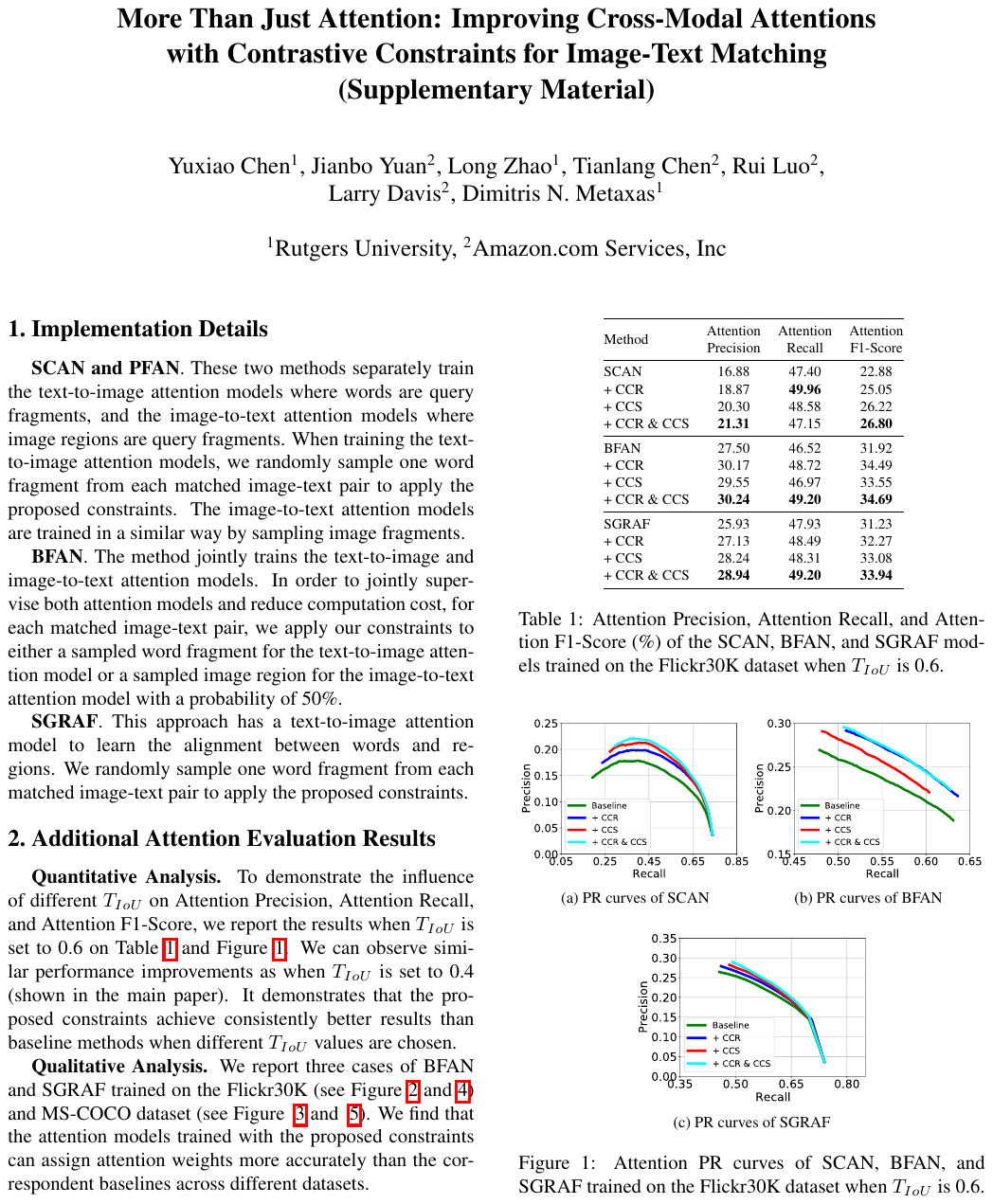}
\includepdf[pages=1-]{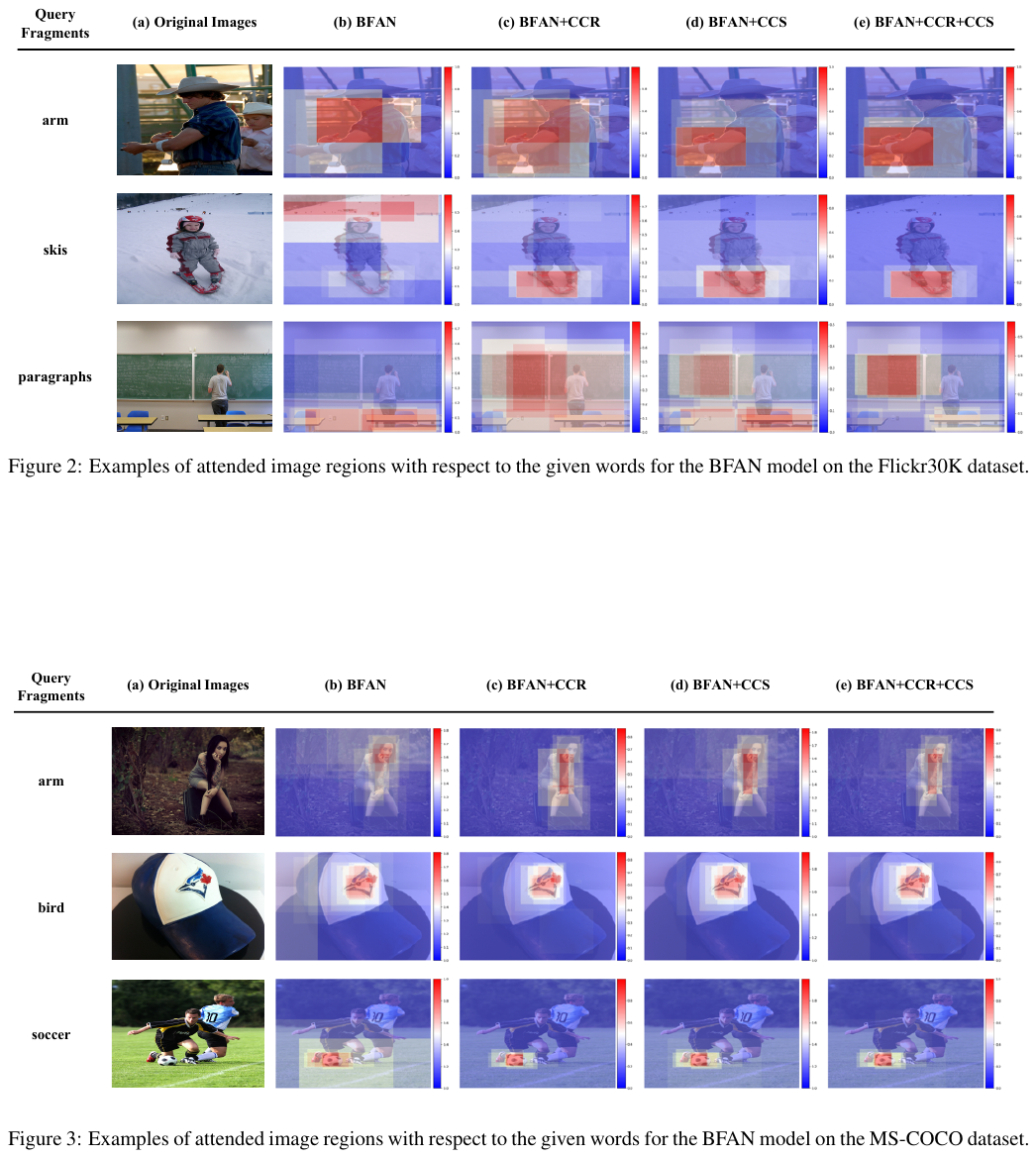}
\includepdf[pages=1-]{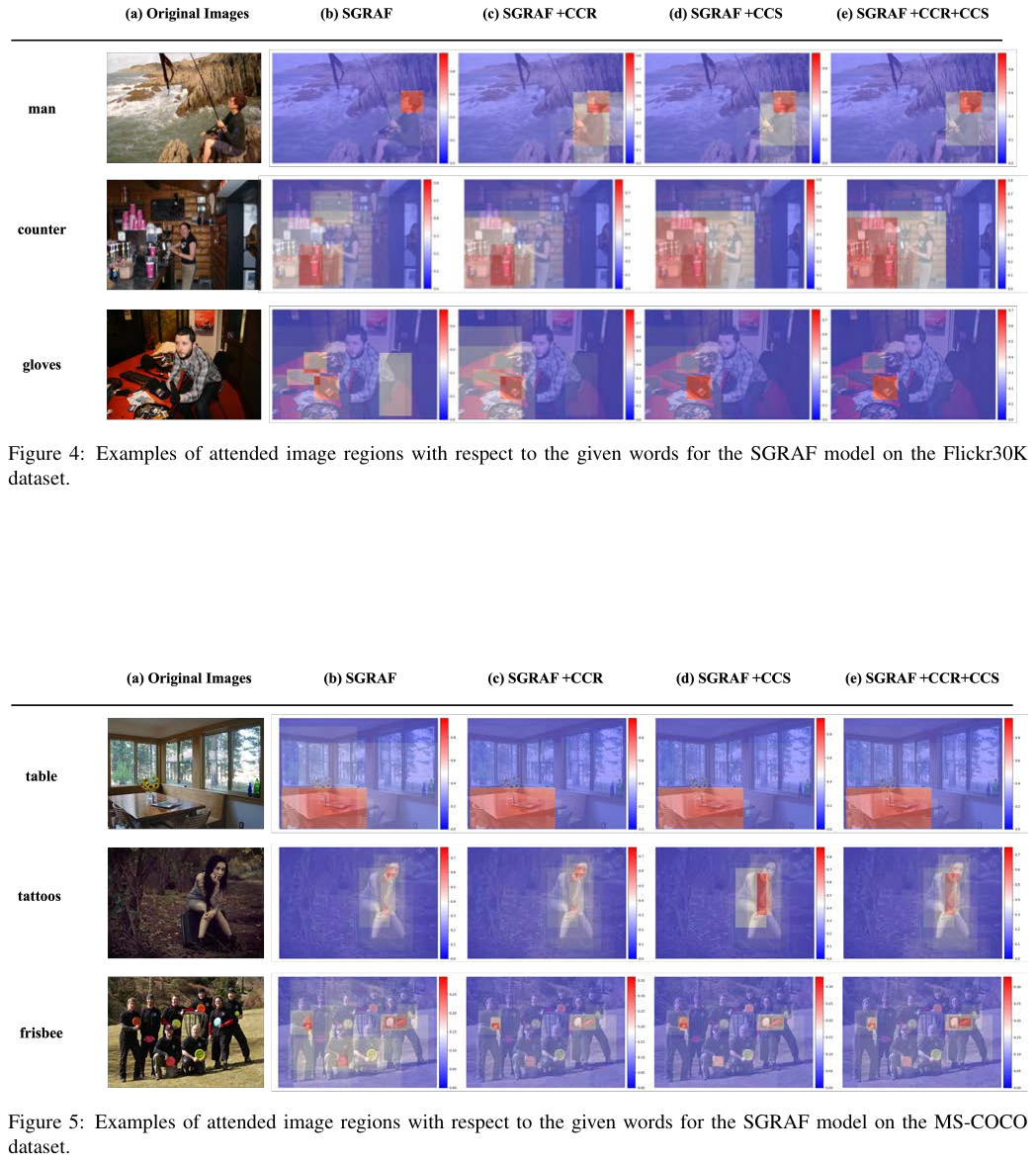}
\end{document}